\documentclass{article}

\usepackage[final]{nips_2017}
\usepackage[utf8]{inputenc} % allow utf-8 input
\usepackage[T1]{fontenc}    % use 8-bit T1 fonts
\usepackage{hyperref}       % hyperlinks
\usepackage{url}            % simple URL typesetting
\usepackage{booktabs}       % professional-quality tables
\usepackage{amsfonts}       % blackboard math symbols
\usepackage{nicefrac}       % compact symbols for 1/2, etc.
\usepackage{microtype}      % microtypography
\usepackage{graphicx}
\usepackage{amsmath}

\usepackage{natbib}

\title{Cross Domain Adaptation using Adversarial networks with Cyclic loss}

\author{
  Manpreet Kaur\\
  CICS, UMass Amherst\\
  \texttt{manpreetkaur@umass.edu} \\
  \And 
  Ankur Tomar\\
  CICS, UMass Amherst\\
  \texttt{atomar@umass.edu} \\
  \And 
  Srijan Mishra\\
  CICS, UMass Amherst\\
  \texttt{smishra@umass.edu} \\
  \And
  Shashwat Verma\\
  \texttt{shashwatverma14@gmail.com} \\
}

\begin{document}

\maketitle

\begin{abstract}
  Deep Learning methods are highly local and sensitive to the domain of data they are trained with. Even a slight deviation from the domain distribution affects prediction accuracy of deep networks significantly. In this work, we have investigated a set of techniques aimed at increasing accuracy of generator networks which perform translation from one domain to the other in an adversarial setting. In particular, we experimented with activations, the encoder-decoder network architectures, and introduced a Loss called cyclic loss to constrain the Generator network so that it learns effective source-target translation. This machine learning problem is motivated by myriad applications that can be derived from domain adaptation networks like generating labeled data from synthetic inputs in an unsupervised fashion, and using these translation network in conjunction with the original domain network to generalize deep learning networks across domains.
\end{abstract}

\section{Introduction}
The problem of input distribution bias has been a long standing problem in deep networks, which is a natural outcome of the MLE framework to arrive at optimized weights. Based on the fact that data collected in real world is often localized in its distribution and entails huge efforts with respect to labeling; the concept of Domains for input space is often used. Domains in an NLP context could refer to the source that the input comes from, for e.g. Wall Street Journal forms one domain of input space while MEDLINE, a medical journal, forms another. In the context of vision problems, we could have MNIST (handwritten digits) as one domain, and SVHN (digits written on real world objects) as another domain. A network trained on a particular task in one domain doesn't give a similar accuracy on other domain. \\

Some interesting problem statements naturally arise from this aspect of deep learning. Is it possible to generalize one single network to generalize well across different domains? Many Machine Learning techniques try to answer this question, for example, auto-encoders which share the same latent space. Other techniques work on an associated problem, which is to learn mappings from one domain(often called source domain) to another(often called target domain) and use a combination of translation and the learned task to generalize better. Solving the domain translation problem also leads to an interesting application apart from improved cross-domain generalization of deep networks, that is, to generate labeled data for a particular domain using this translation in an unsupervised fashion. In what follows for this report, we define the following terms. Source domain refers to the domain of input space where we have labels and a corresponding task that we want to generalize for the target domain. Target domain refers to the domain of input space which lacks any labels.\\

While a lot of development has happened in this field, the class of methods known as Domain Adaptation suffer from the problems associated with any Adversarial training, as the cross domain translation networks are trained using a GAN system. Therefore, we suffer, chiefly amongst many problems, from an objective evaluation of the performance of GAN system. Many such GAN frameworks don't work on domain translation, but instead learn a common latent representation for the two domains. Yet other methods work by sharing parameters to learn an effective mapping and task performance simultaneously. Even though GAN frameworks for training translation systems force the translation from source domain to target domain to have the similar distribution, they fail to impose the vis-a-vis mapping for the task on source domain we are trying to generalize. For example, consider MNIST learning task, which is to classify the digits. A GAN based adaptation method might be successful in translating SVHN images to something that resembles MNIST digit image distribution, but the framework doesn't guarantee that a 7 from MNIST is mapped to a 7 from SVHN database, it could easily be mapped to a 2 or a 3. To meet the order preserving translation to generalize a regression task is a significantly more difficult problem.\\

In this work, we propose several modifications to GAN training routine and introduce a new loss in context of domain adaptation so that the domain translation is order preserving. The two domains we looked at are real world and synthetic self driving datasets. The real world camera images come from Comma.ai dataset and the synthetic video-game like images are from the Udacity dataset. We tried to learn a steering angle prediction task which generalizes well across both the domains, by transferring the real world comma.ai dataset into the synthetic domain of inputs, for which we have steering angle labels. We have a set of 4 networks which we are training in a phase-wise manner. A domain translation network $G_{S\rightarrow T}$ performs the task of synthesizing target domain from source images. Another domain translation network $G_{T \rightarrow S}$ synthesizes Source images from target images and forms a complement to the first $S\rightarrow T$ network and is important for the cyclic Loss training which we explain in detail in later sections. We have two Discriminator networks , $D_{S \rightarrow T}$ and $D_{T \rightarrow S}$  which basically classifies the input image into source or target domain. The fifth network we have in our whole process is the steering angle prediction network $R_{Steering}$.\\

With our goal to finally achieve a good accuracy on the combined network $G_{T \rightarrow S}$ in conjunction with the steering angle prediction network $R_{Steering}$, we divided our training process in 3 phases. First, we trained the network $R_{Steering}$ to achieve good generalization on the source input domain. In the second phase, we trained the $G_{S \rightarrow T}$ with $D_{S \rightarrow T}$ and $G_{T \rightarrow S}$ with $D_{T \rightarrow S}$ independently under a GAN setting. In the third and final phase, we combine all the working pieces, and do a composite training of all the networks simultaneously starting with their pre-trained states from phase 1 and phase 2 based on a set of losses. In terms of results, we faced roadblocks in phase 2 with GAN training, but we achieved a synthesis from target to source which faithfully translated the high level semantic features. In terms of the final generalized regression network $R_{Steering}$, we saw a reduction of Loss by 0.13 and an increase of 12.09\% in accuracy (AARE, Average Absolute Relative Error) on hold out validation set while predicting from source data synthesized from target data, using phase 3 training as opposed to predicting from Phase 1 trained regression network on source data itself. 
\section{Related Work}

Convolutional neural networks when trained on large-scale datasets, act as very powerful tools which can be used across a wide variety of tasks and visual domains. However, sometimes due to the phenomenon of dataset bias or domain shift the models which are trained along with these representations do not generalise well when they see new dataset. This problem can be solved in many different ways. The the group of models that are used for such applications are called Transfer Learning. In order to go ahead and define the different kinds of related work that has been done in this field, we would like to define transfer learning in a detailed way. Transfer learning or inductive transfer is a research problem in machine learning that focuses on storing knowledge gained while solving one problem and applying it to a different but related problem. Given source and target domains $D_s$ and $D_t$ where $D={X,P(X)}$ and source and target tasks $T_s$ and $T_t$ where $T = {Y,P(Y|X)}$ there are four ways in which the scenarios can differ:

1) $X_s \neq X_t$: This happens when the source and the target differ in feature spaces. There has been a lot of work in this field mostly in the field of Natural Language Processing. This particular field is reffered to as cross-lingual adaptation. 

In \citep{prettenhofer2010cross}, used cross-lingual adaptation for transfer learning between different languages. They defined a method called Structural Correspondance Learning in which they used a word translation oracle to transfer classified data from one language to another. Further they also undertook experiments in different languages assigning English as the primary language and did sentiment analysis for German and French and got cutting edge results.

2) $Y_s \neq Y_t$: This happens when the label spaces are different between the two tasks. This occurs mostly with scenario 3 and hence we have given a better review in that point.

3) $P(Y_s|X_s) \neq P(Y_t|X_t)$: The conditional probability distributions are not the same for the source and the target task. This finds a lot of application in tasks of over-sampling and under-sampling and also things like SMOTE analysis. 

4) $P(X_s) \neq P(X_t)$: This is the case when the marginal probability distributions in the source and target domains are different. This scenario is called Domain Adaptation. This is the sub-category of Transfer Learning in which most of our literature review and hence our work has been in. The initial work in this direction was through the usage of certain class of methods which use Maximum Mean Discrepancy(MMD) loss to calculate the difference between two domain means-source and target. 

\citep{gretton2009covariate} achieved this by matching covariate distributions in high dimensional feature spaces. This method did not require distribution estimation as sample weights are obtained by quadratic programming. This method could deal very well with simple covariate shifts but did not work for large scale deviations between the source and target. 

\citep{tzeng2014deep} used Deep Domain Confusion(DDC) for maximizing domain invariance. The DDC method coupled with Maximum Mean Discrepancy was used along with classification loss which made it both discriminative and domain invariant. This model used a CNN model with an adaptation layer which when coupled with a domain confusion loss gives a representation that is domain invariant. 

\citep{long2015learning} applied Maximum Mean Discrepancy to layers embedded in a kernel Hilbert space matching higher order statistics of the two distributions. They called this Deep Adaptation Network. The problem that the Deep Adaptation Network overcame was that of lack of labelled training data for the target domain and hence it reduced the problem of overfitting. It was a remarkable breakthrough in this class of networks and hence the Domain Adaptation task as well. In this paper, they extended the AlexNet model to achieve their objective. 

In \citep{tzeng2015simultaneous} they proposed a CNN architecture for sparsely labeled target domain data. Meanwhile, they also optimized for domain invariance which facilitates domain transfer and used a soft label distribution which matches loss to inter-task transfer information. They also transferred the similarity structure amongst categories from source to target to optimize the produced representation in the target domain by using few target labeled examples as references. To do so, an average output probability distribution was computed and then and a direct optimization to match the class distribution was done for each labeled example. This lead to a task adaptation task through information transfer to categories with no explicit labels in the target domain.

In \citep{ganin2015unsupervised} proposed a gradient reversal algorithm, also known as ReverseGrad which treated domain invariance as a binary classification problem and directly maximised the loss of the domain classifier by reversing its gradients. As the training progresses, the approach generates deep features that are discriminative in the main source domain and also invariant with respect to the domain shift. This is achievable in all feed-forward models through addition of standard layers and a gradient reversal layer. The resulting augmented architecture is trained using backpropagation.

In \citep{goodfellow2014generative} used a Generative model that pits two models against each other. One of the models generates fake images while the other gives true images and the Discriminator tries to maximise its payoff by predicting the correct label for a particular example being fake or real. We end up optimizing a minimax loss to a point where the discriminator is no longer able to differentiate between the real and fake images. 

In \citep{tzeng2017adversarial} the authors outlined a generalized framework for adversarial adaptation which used discriminative modeling, untied weight sharing and a GAN loss which they called Adversarial Discriminative Domain Adaptation (ADDA). In this, they learn a discriminative mapping of target images to the source feature space which aims to fool a domain discrimator whose objective is to distinguish encoded target images from source examples.

Finally, in \citep{hoffman2018cycada}, the authors have introduced a cycle consistent loss function, in which they learn two generator functions to adapt domain from source to target data distribution, and target to source data distribution. The source data is passed through both generators sequentially, and an L2 loss between original source image and reconstructed source image is used to train both generator networks in an adversarial style. 

We take inspiration from all of the above cited works and apply domain adaptation to our task of steering angle prediction from images.

\section{Methodology}
Our experiments are based on modifying aspects of GAN training which would allow the discriminator to not win excessively and render the Generator network mute. Along with that, we proposed certain modifications and used a symmetric GAN network which performs a translation from source to target domain and reinforces the Generator networks to preserve ordinality of target to source transformation on the regression task. We define below the architectures for different component networks of our composite architecture, and then move on to describe phase-wise the methodology to train these networks.
\subsection{Component Networks}
We have 5 networks which we are training to learn effective Target to source transformations. 

\begin{table}[t]
  \caption{Network architecture details for all networks used in our framework}
  \label{sample-table}
  \centering
  \begin{tabular}{ll|ll|ll}
    \toprule
    \multicolumn{2}{c}{$G_{S \rightarrow T}, G_{T \rightarrow S}$} & \multicolumn{2}{c}{$D_{S \rightarrow T}, D_{T \rightarrow S}$} & \multicolumn{2}{c}{$R_{Steering}$}                   \\
    Layer     & Specs& Layer     & Specs& Layer     & Specs \\
    \midrule
    Conv Layer 1 & 24 Channels & Conv Layer 1 & 24 Channels & Conv Layer 1  & 24 Channels   \\
    & 5X5 Window && 5X5 Window && 5X5 Window      \\
    Maxpool 1 & 2X2 Window & Maxpool 1 & 2X2 Window & Maxpool 1 & 2X2 Window  \\
    & 2 stride && 2 stride && 2 stride      \\
    Conv Layer 2 & 48 Channels & Conv Layer 2 & 48 Channels & Conv Layer 2  & 48 Channels   \\
    & 5X5 Window && 5X5 Window && 5X5 Window      \\
Maxpool 2 & 2X2 Window & Maxpool 2 & 2X2 Window & Maxpool 2 & 2X2 Window  \\
    & 2 stride && 2 stride && 2 stride      \\
    Conv Layer 3 & 64 Channels & Conv Layer 3 & 64 Channels & Conv Layer 3  & 64 Channels   \\
    & 3X3 Window && 3X3 Window && 3X3 Window      \\
  Maxpool 3 & 2X2 Window & Maxpool 3 & 2X2 Window & Maxpool 3 & 2X2 Window  \\
    & 1 stride && 1 stride && 1 stride      \\
    
    Unpool 4 & 2X2 Window & FC1 & 100 units & FC & 1 unit  \\
    & 1 stride &&&&\\
   
   Deconv Layer 4 & 48 Channels & Final FC layer & 2 units  \\
    & 3X3 Window       \\
    
     Unpool 5 & 2X2 Window \\
    & 2 stride      \\
    Deconv Layer 5 & 24 Channels \\
    & 5X5 Window       \\
    
     Unpool 6 & 2X2 Window \\
    & 2 stride      \\
    
    Deconv Layer 6 & 3 Channels \\
    & 3X3 Window       \\
  
    \bottomrule
  \end{tabular}
\end{table}

\subsubsection{Steering Regression Network $R_{Steering}$}
This network is responsible for the core learning task, which is to predict the steering angle given an image. To perform this, we built a convolutional neural network with 3 convolution block, and one fully connected block. The non-linearities we used were ReLU and we used the least squares loss for regression output. Each convolution was batch normalized, and each block down-sampled the image by a factor of half using Max Pooling. Our input size is 80X160 pixels with 3 channels. The convolution layers have 5X5, 5X5, and 3X3 windows, with 24, 48 and 64 channels after their outputs respectively and are batch normalized. These blocks are followed by a linear layer with one output for the steering angle. 
\begin{figure}
\centering
\includegraphics[width=1.0\textwidth]{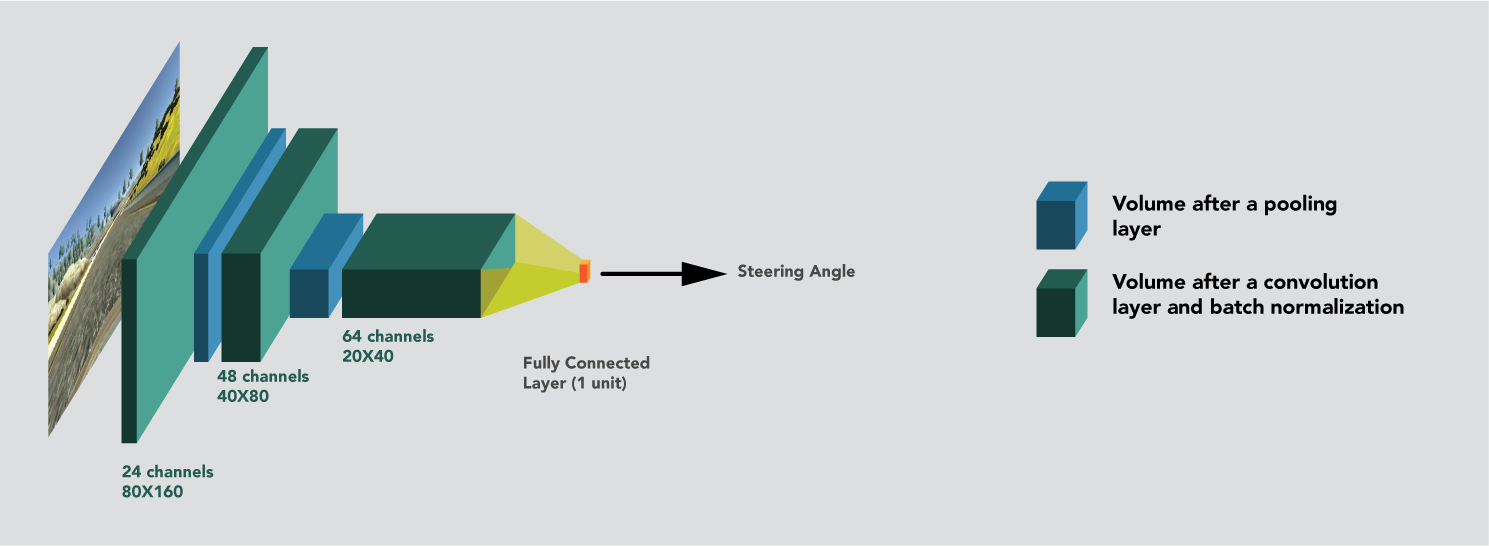}
\caption{Network used for Steering Angle Prediction $R_{Steering}$}
\end{figure}

\subsubsection{Source to Target and Target to Source Domain Translation networks $G_{S \rightarrow T}$ \& $G_{T \rightarrow S}$ }

This domain translation network is inspired by autoencoder architectures, mainly based on the hypothesis that such architectures would force the network to distill latent features that are semantically important from which the source domain image can be synthesized. Specifically, the domain translation network consists of 6 layer blocks. The first 3 layers perform convolution operations and the later 3 perform deconvolution operations along with un-pooling. The image while passing through these layers has 24, 48 , 64, 48, 24, and finally 3 channels respectively. The window sizes vary as 3X3, 3X3, 5X5, 5X5, 3X3, and 3X3 through the layers. Each block is batch normalized, and two Max Pooling and Unpooling schemes have been used. One is a non-overlapping 2 stride, 2X2 window Max Pooling, and other is the 1 stride, 2X2 window Max Pooling.\\
\begin{figure}
\centering
\includegraphics[width=1.0\textwidth]{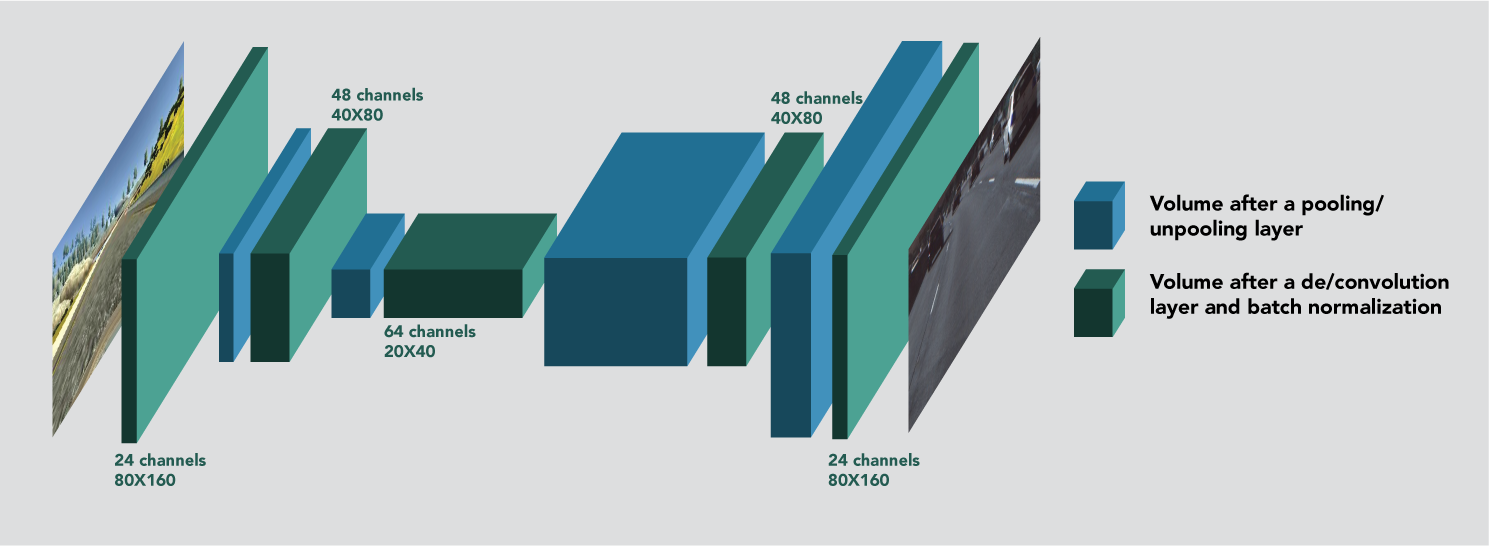}
\caption{Schematics for the domain translation network $G_{S \rightarrow T}$
}
\end{figure}

We also considered adding skip connections between the second and fourth layers, but decided against it as experimentation was taking a long time. All layers had a leaky ReLU to improve GAN training and prevent gradient vanishing.
\subsubsection{Discriminator Networks $D_{ST}$ \& $D_{TS}$}
This network essentially mimicked the encoder part of the auto-encoder inspiration we took for the domain translation network and consisted of 3 convolution layer blocks with Leaky ReLUs. These convolutional layers were followed by two fully connected layers. The inner FC layer had 100 units, and the final FC layer had two units, each representing a domain.\\

\begin{figure}
\centering
\includegraphics[width=1.0\textwidth]{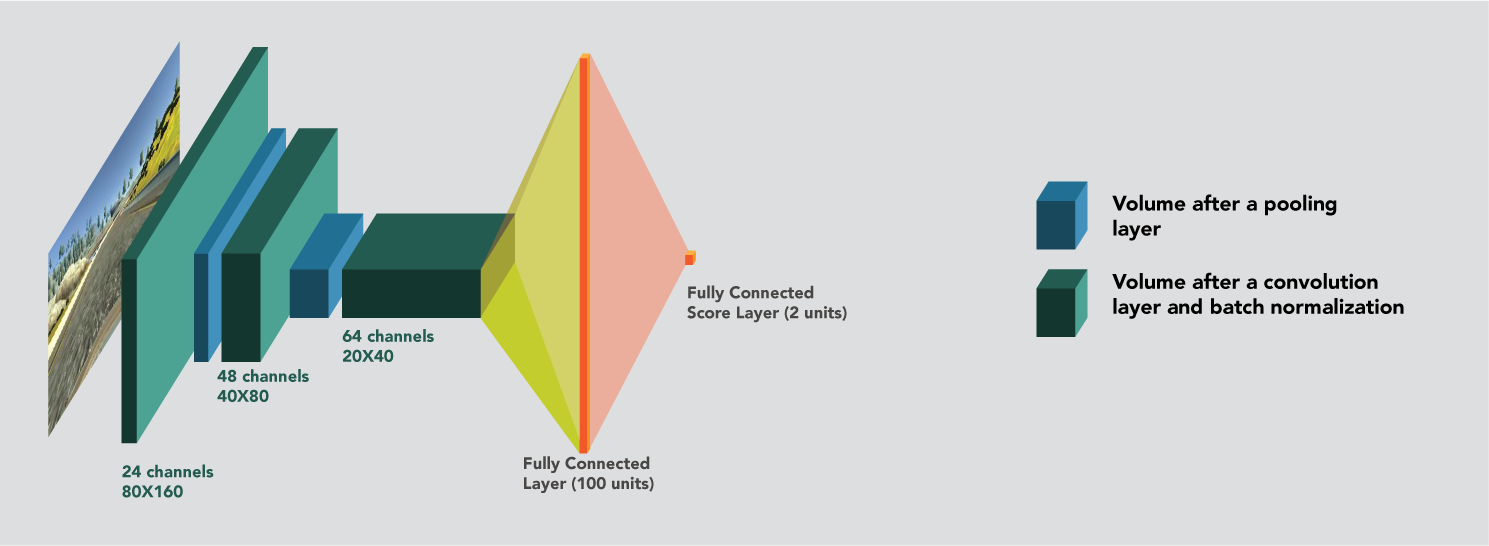}
\caption{Schematics for the discriminator network $D_{S \rightarrow T}$
}
\end{figure}

The architectures for all these networks are explained in detail in Table 1. We next elaborate on our loss functions, modifications specific to GAN training and training routines.

\subsection{Loss Functions and Overall Training Methodology}
In this section, we define the overarching process we followed to ensure our Domain Translation networks had robust learning, and this section entails the most important components of our proposed hypotheses, the reconstruction loss. In essence, we are going to train simultaneously all the networks in this setup using a composite that concurrently tries to align all the goals for our different learning tasks. As initialization to this process, we will use pretrained individual networks $G_{S \rightarrow T}$, $G_{T \rightarrow S}$ and $R_{Steering}$. Therefore, our process is a 3-phase method, where Phase 1 trains a regression network on Source Domain (the Udacity Dataset) and Phase 2 trains the GAN setup networks. In the final phase 3, we train all the 5 networks in a concurrent fashion to achieve goals important for all the tasks that are taking place in each of these networks. We describe in detail the Learning framework used for each of the 3 phases.\\
\subsubsection{Phase 1 : Training Steering Angle Regression}
This 4 layer network was trained using the Empirical Risk Minimization framework, which translated to minimizing the mean squared error for predicted steering angle. We used the loss\\
\begin{equation} \label{eq1}
\begin{split}
\mathcal{L}_{Steering}(X_s,Y_s,f_s) &= \mathbb{E}_{x_s \sim X_S}[||f_s(x_s)-y_s ||_2]\\ 
& =(1/n)\sum_{i=1}^n(f_s(x_i)-y_i)^2
\end{split}
\end{equation}

This was a fairly standard network to train, and we used Xavier Initialization for weight initializations, a learning rate of 1e-4 and second order Adam optimizer for optimization.

\subsubsection{Phase 2 : Training the Domain Translation and Discriminators in GAN setting}

This part of the training process required significant amount of engineering which was not driven strictly by strong mathematical background. Our Loss for the domain translation network $G_{S \rightarrow T}$ and $D_{S \rightarrow T}$ was\\
\begin{equation} \label{eq2}
\begin{split}
\mathcal{L}_{GAN}(G_{S \rightarrow T}, D_{S \rightarrow T}, X_S,X_T)=  & \mathbb{E}_{x_t \sim X_T}[log(D_{S \rightarrow T}(x_t)]\\
& +\mathbb{E}_{x_s \sim X_S}[log(1-D_{S \rightarrow T}(G_{S \rightarrow T}(x_s))]
\end{split}
\end{equation}
This is a conventional 2 player based min-max zeros sum game loss. Even though its possible in theory and has been demonstrated in practice to show effective results, the training process for this min-max loss often turns out to be unstable and fails to converge. Another aspect to consider is that despite ensuring that $G_{S \rightarrow T(x_s)}$ will follow a distribution similar to $X_T$, we never know whether the objective will enforce any structural or content retention from the $x_s$ samples. We will try to address this problem in the third phase, and instead, here, we try to solve the optimization problem of $\max_{G_{S \rightarrow T}}\min_{D_{S \rightarrow T}}\mathcal{L}_{GAN}$ so that we can use these as pre-trained networks in phase 3. We do the exact same training process to train the networks $G_{T \rightarrow S}$ and $D_{T \rightarrow S}$. In particular, we first employed Leaky ReLUs to prevent any of the units from suffering with vanishing gradients. Secondly, we utilized a threshold based training routine to alternate between training of the Generator and Discriminator network. Thirdly, in our minibatches for training, we alternated between full source minibatch and full target minibatch instead of a mixed training minibatch consisting of both synthesized target images, and original target samples.\\

\subsubsection{Phase 3 : Training all the networks together}

This part of the training entailed loading the networks trained in previous two phases as pretrained networks for the third phase. To enforce semantic retention, it should be the case that $G_{T \rightarrow S}(G_{S \rightarrow T}(x_s)) \sim x_s $, and similarly $G_{S \rightarrow T}(G_{T \rightarrow S}(x_t)) \sim x_t $. While we want to keep on refining the $G_{S \rightarrow T}$, $G_{T \rightarrow S}$, $D_{S \rightarrow T}$, $D_{T \rightarrow S}$, and $R_{Steering}$ simultaneously so that they achieve their individual goals we would like to individual pieces of this network group to reinforce (or optimally play against) each other. One way of doing this is to use the idea mentioned above, and introduce a loss which reinforces this reconstruction aspect of the domain translation networks. We will call this the reconstruction loss, i.e. \\
\begin{equation} \label{eq3}
\begin{split}
\mathcal{L}_{Rec}(G_{S \rightarrow T}, G_{T \rightarrow S}, X_S, X_T) = & \mathbb{E}_{x_s \sim X_S}[||G_{T \rightarrow S}(G_{S \rightarrow T (x_s))}-x_s||_1]\\
& +
\mathbb{E}_{x_t \sim X_T}[||G_{S \rightarrow T}(G_{T \rightarrow S (x_t))}-x_t||_1]
\end{split}
\end{equation}

Another way we are trying to enforce the semantic retention in the domain translation networks is by training the regression task network to predict from both the source images, as well as the generated images from these source images, labeled using the source label. Therefore, our objective function looked like\\
\begin{equation} \label{eq4}
\begin{split}
\mathcal{L}_{combined} & (X_S,X_T,Y_S,G_{S \rightarrow T},G_{T \rightarrow S},D_{S \rightarrow T},D_{T \rightarrow S},R_{Steering})=\\
& \mathcal{L}_{GAN}(G_{T\rightarrow S},D_{T \rightarrow S}, X_S,X_T) +\\
& \mathcal{L}_{GAN}(G_{T\rightarrow S},D_{T \rightarrow S}, X_S,X_T) + \mathcal{L}_{Rec}(G_{S\rightarrow T},G_{T \rightarrow S}, X_S,X_T) +\\
& \mathcal{L}_{Regression}(f_S,X_S,X_T,Y_S,G_{S \rightarrow T})
\end{split}
\end{equation}

With parameters shared between different losses which form $\mathcal{L}_{combined}$, our expectation was a more stable training of the GAN setting and a more generalized $R_{Steering}$. However, this was a very expensive and tricky training process with many possibilities for sequential training, much like the sequential training often followed in GANs. In essence, our objective optimization problem was\\
\begin{equation} \label{eq5}
\begin{split}
S^*=\min_{f_S}\min_{\substack{G_{S \rightarrow T}\\G_{T \rightarrow S}}}\max_{\substack{D_{S \rightarrow T} \\ D_{T \rightarrow S}}}(L_{Combined})
\end{split}
\end{equation}

We implemented the gradient step for this optimization problem in modular manner, much similar to a single GAN training, where we use a complement of the objective function to solve for the max problem. This forces the training to cycle between different data pieces and networks, and reduces the chance of going in the true gradient direction of the objective function, which in itself is not a guarantee on reaching the saddle point we wish to find for $L_{Combined}$. So hoping for the best, we took mini-batches which cycled through 1. Source Images, 2. Synthesized Source Images from Target Images, 3. Target Images, and 4. Synthesized target images from Source Images, where we updated the steering angle regression only during the mini-batches 1 and 4 (attaching the source label while synthesizing from them).\\

With this complicated scheme of training, we tried another idea for forcing semantic robustness and stable GAN training by making the task for the Discriminator harder. In the earlier schemes, $D_{S \rightarrow T}$ basically solved the task of "whether an input is from target domain or not". Similarly, $D_{T \rightarrow S}$ solves the task of "whether an input image is from source domain or not".
 For the sake of our framework, where we have both $G_{S \rightarrow T}$ and $G_{T \rightarrow S}$, it might be more helpful to have a general Discriminator $D$, whose task is to classify the image as source or target. Therefore, if we replace all the discriminators in our previous scheme with this "general" discriminator, our hypothesis is that in GAN training method with shared discriminator for the cyclic domain translators, the shared discriminator will have to learn a significantly harder task and thus would be less prone to unstable winning in the zero-sum game. This woul reduce our networks from 5 to 4, and the optimization problem will look like:\\

\begin{equation} \label{eq5}
\begin{split}
S^*=\min_{f_S}\min_{\substack{G_{S \rightarrow T}\\G_{T \rightarrow S}}}\max_{D}(L_{Combined})
\end{split}
\end{equation}

The schematics for both the shared and distinct discriminators are shown in the figure.

\begin{figure}
\centering
\includegraphics[width=1.0\textwidth]{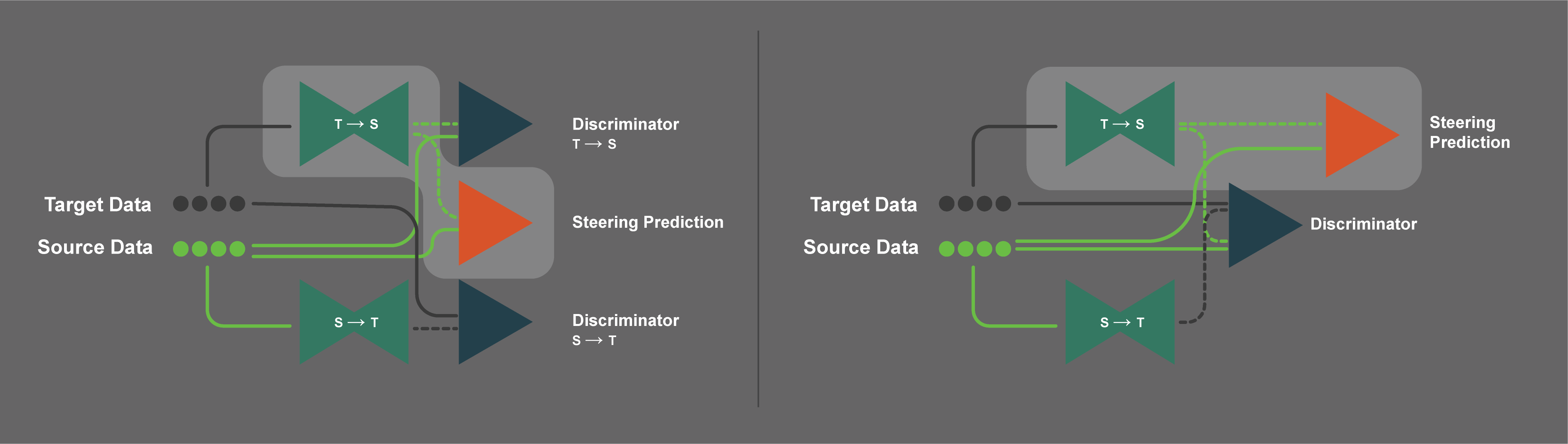}
\caption{Schematics for the overall phase 3 training. Left side shows the schema with 5 networks with separate discriminators. Right hand side shows a network structure with a unified discriminator; a hypothesis we weren't fully able to test.
}
\end{figure}

\section{Datasets}

1) Dataset for which we have trained the labels(\textbf{source})- Udacity self driving dataset\\
2) Dataset for which we have done the domain adaptation- comma.ai real world dataset(\textbf{target})\\

\textbf{Udacity self-driving data:}\\
The Udacity self-driving data was created for there Machine Learning nanodegree program. It was done in collaboration with Google and the objective was to initiate more research in the space of self-driving cars. The dataset consists of 8032 images which are extracted from a simulator in a gameplay environment. The download link is: \url{https://github.com/ymshao/End-to-End-Learning-for-Self-Driving-Cars}. Some of the images of the dataset are:\\
\begin{figure}[htbp]
\centering
\includegraphics[width=0.4\textwidth]{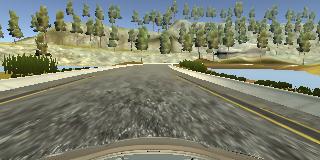}
\includegraphics[width=0.4\textwidth]{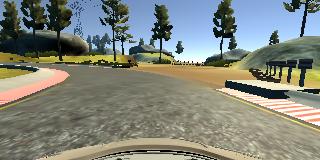}
\caption{Some images from the Udacity dataset}
\end{figure}

The images have been generated through the simulator that Udacity has created and images can be generated by any run of the simulator which simulates it like a driving environment and the vehicle can be controlled through the four arrow keys on the keyboard(like any other simple game). The dataset gives different values of throttle, driving angle through 3 different camera views i.e. centre, left and right. We have chosen the central frame and have used the images in size $80 X 160$. We chose this dataset because of the ease at which any number of examples can be generated with high amount details concerning the steering angle. In comparison to this, any physical simulator would require  physical measurement equipment to measure steering angle and also a different pipeline to translate throttle data to computer comprehensible data. Also, would require a US driving license which none of the team members had(:P).\\

\textbf{Comma.ai self-driving data:}\\

The comma.ai dataset is a real-world dataset that is also made by central camera mount over the driver's dashboard. It is also publicly available for academic use and is intended to enhance research in the self-driving space. Comma.ai is a company which was actively involved in the space of self-driving cars. Some of the sample images of the comma.ai dataset are as follows. Also, note that the dataset consists of images both during the day and the night time.
\begin{figure}[htbp]
\centering
\includegraphics[width=0.4\textwidth]{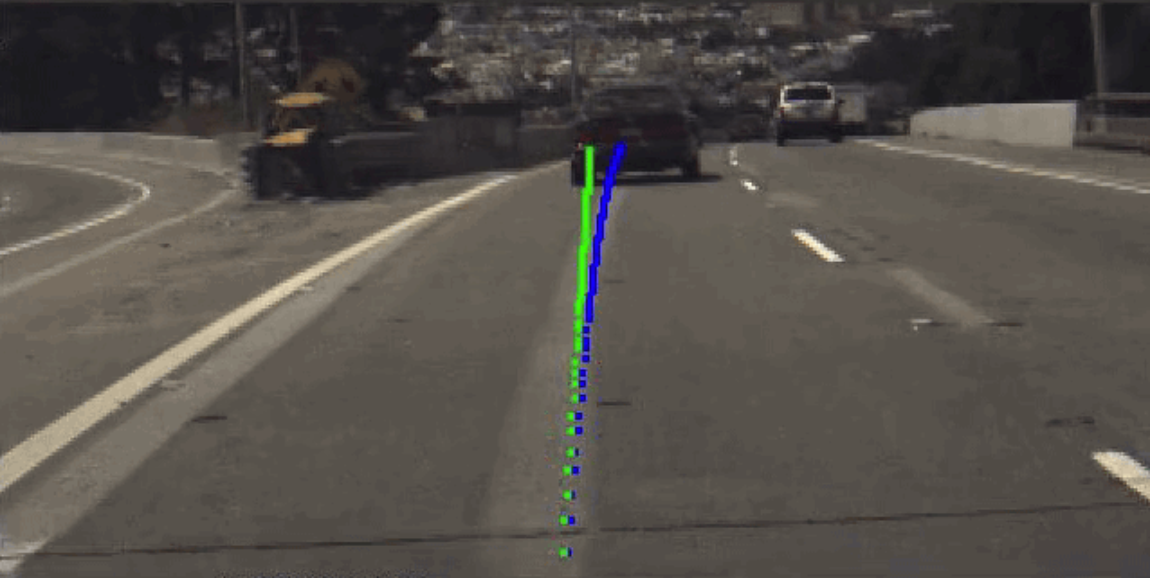}
\includegraphics[width=0.4\textwidth]{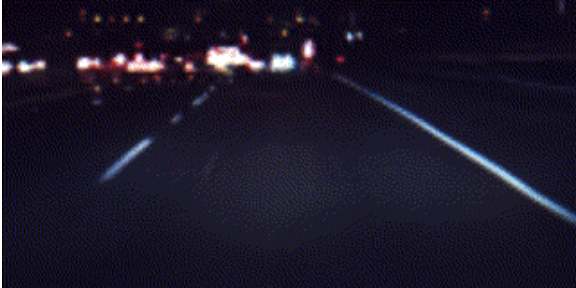}
\caption{Some images from the comma.ai dataset}
\end{figure}

The images that we are using for the comma.ai dataset are also $80X160$ and we have done the task of domain adaptation and hence have used the labels of the dataset only to see the test accuracy. The images are 7 and a half hours of highway driving videos converted to videos and hence it is a lot of data for any good machine learning task. The reason for choosing the dataset is that it makes the task very valuable in terms of the kind of objectives that can be achieved through it. The link to the dataset is: \url{https://github.com/commaai/research}.

\section{Experiments}
We build our network as described in the methodology. The training process takes place in three phases. The first phase trains the convolution network on source data. This is a fully supervised task as we train source images with their steering angle labels. The second phase trains the domain translation networks and the corresponding discriminators using GAN loss and reconstruction loss. The third phase uses pre-trained networks from Phase 1 and 2 with the final aim of learning $R_{Steering}$ for both source and target domains. In the third phase, all the network parameters of $G_{S\rightarrow T}$, $G_{T\rightarrow S}$, $D_{S\rightarrow T}$, $D_{T\rightarrow S}$, $R_{Steering}$ are trained using a composite loss of GAN loss, reconstruction loss and semantic loss. We now describe the hypothesis we formed to train our networks and the experiments designed to validate or refute them. The discriminators are binary classification networks which try to distinguish the input between source and target domains. The discriminators minimize their cross-entropy loss $L_{discrim}$ while generators try to maximize the discriminator loss. \\
\\
\subsection{Phase 1: Training the regressor}
We used a convolutional neural network to train the regressor. There were multiple experimentations both on the total number of layers used as well as the normalization of the image.\\

{\bf Hypothesis 0: Using two convolutional layers w/o normalization}
Using two convolutional layers the loss decrease was very slow and could barely be witnessed. After this experimentation we felt there was a need to increase the number of convolutional layers.

{\bf Hypothesis 1: Using three convolutional layers w/o normalization}
Using three convolutional layers considerably improved the loss plots. Till this point we had not used normalised images and wanted to experiment with the same to test the losses downstream on the generators.

{\bf Hypothesis 2: Using three convolutional layers with normalization}
Normalizing the images helps reduce the loss of the generator much more than that without normalization. This seems to work better with the whole pipeline and hence was finalized in the whole network architecture.

\subsection{Phase 2: Training domain translation network}
We use Adversarial learning approach to train $G_{S\rightarrow T}$, $G_{T\rightarrow S}$. We introduce two discriminator networks $D_{S\rightarrow T}$ and $D_{T\rightarrow S}$ which will be trained adversarially with respect to $G_{S\rightarrow T}$, $G_{T\rightarrow S}$. 

{\bf Hypothesis 0: } In all our experiments described below, we normalized the data using mean and standard deviation of 128.0, 47.0 and 70.0, 44.6 for source and target data respectively. The mean and std deviation values were calculated over all the channels of training data. This decision was inspired by works of \citep{norm}, \citep{DCGAN}, etc.

{\bf Hypothesis 1: Leaky ReLu vs ReLu} \\
We hypothize that leaky ReLu will bring more stability to adversarial training as non-zero gradients flow backwards. To test this hypothesis, we plot the losses of generator and discriminator and check if adding leaky ReLu stabilizes learning.

{\bf Hypothesis 2: } Theoretically, maximizing the loss of discriminator is equivalent to minimizing $log(1-D_{S\rightarrow T}(data))$. We tried the two different ways of writing objective function for generator network and plotted the loss functions $L_{S\rightarrow T}^{G}$ and $L_{S\rightarrow T}^{D}$ to check which works better empirically.  

{\bf Hypothesis 3: Shuffled mini-batch vs separate mini-batch} \\
The adversarial learning can be performed in iterations i.e. the source and target data is passed over the network iteratively in separate mini-batches or by shuffling source and target data in the same mini-batch. We describe the results of this experiment in the results section. For this experiment, we have compared the convergence/divergence statistics of the loss functions i.e. $L_{S\rightarrow T}^{G}$ and $L_{S\rightarrow T}^{D}$ of both methods over 200 iterations.

{\bf Hypothesis 4: Balanced training or sparse training} \\
Many GAN works done previously have noted that GANs are notoriously hard to train. As noted by \citep{DCGAN}, discriminator tends to win in this mini-max adversarial game as it performs an easier task of binary classification. Hence, we should give some advantage to the generator like updating generator weights more frequently. We used the following algorithm to train our adversarial network.

{\bf Algorithm}
for i $\in$ \{0..epochs\}
\begin{enumerate}
\item while $L_{S \rightarrow T}^{D} \geq threshold $\\
\hphantom{forvj} batch $\leftarrow get\_batch(source)$ \\
\hphantom{forvj} $optimize(batch, L_{S \rightarrow T}^{D})$
\item while $L_{S \rightarrow T}^{G} \geq threshold $\\
\hphantom{forvj} batch $\leftarrow get\_batch(target)$ \\
\hphantom{forvj} If $L_{S \rightarrow T}^{D}) \geq threshold$ \\
\hphantom{forvjforvj} $optimize(batch, L_{S \rightarrow T}^{D})$ \\
\hphantom{forvj} $optimize(batch, L_{S \rightarrow T}^{G})$ \\
\end{enumerate}

Once again, we compare the loss functions of two methods to decide which technique works better.

{\bf Hypothesis 5: GAN loss augmented with Reconstruction loss} \\
GAN training is unstable and it is difficult to generate images that exactly mimic the target or source distribution. Hence, to generate better images, we also use reconstruction loss alongwith adversarial loss to train the network. For this section, we compare the images generated by these two training methods. Both training methods use the best combinations of parameters found in previous hypothesis.

\subsection{Phase 3: Training the regressor} 
The final regressor is trained using all components of the network. We want to generalize the source regressor trained in first phase, over  both source and target domains. Note once again, that we don't use labels from target data. We transform source to target distribution and feed generated target image to the network so that it learns to output correct steering angles for target images. Similarly, we reconstruct the source image from generated target image to preserve semantic labels of the data. Overall, we train the $G_{S \rightarrow T}$, $G_{T \rightarrow S}$ and the regressor $R$ in the final stage of training. We use the composite loss function of adversarial losses, reconstruction losses and the MSELoss. The results are discussed in Results section.

\section{Results}
\subsection{Results for phase 1}
\begin{figure}[htbp]
\centering
\includegraphics[width=0.4\textwidth]{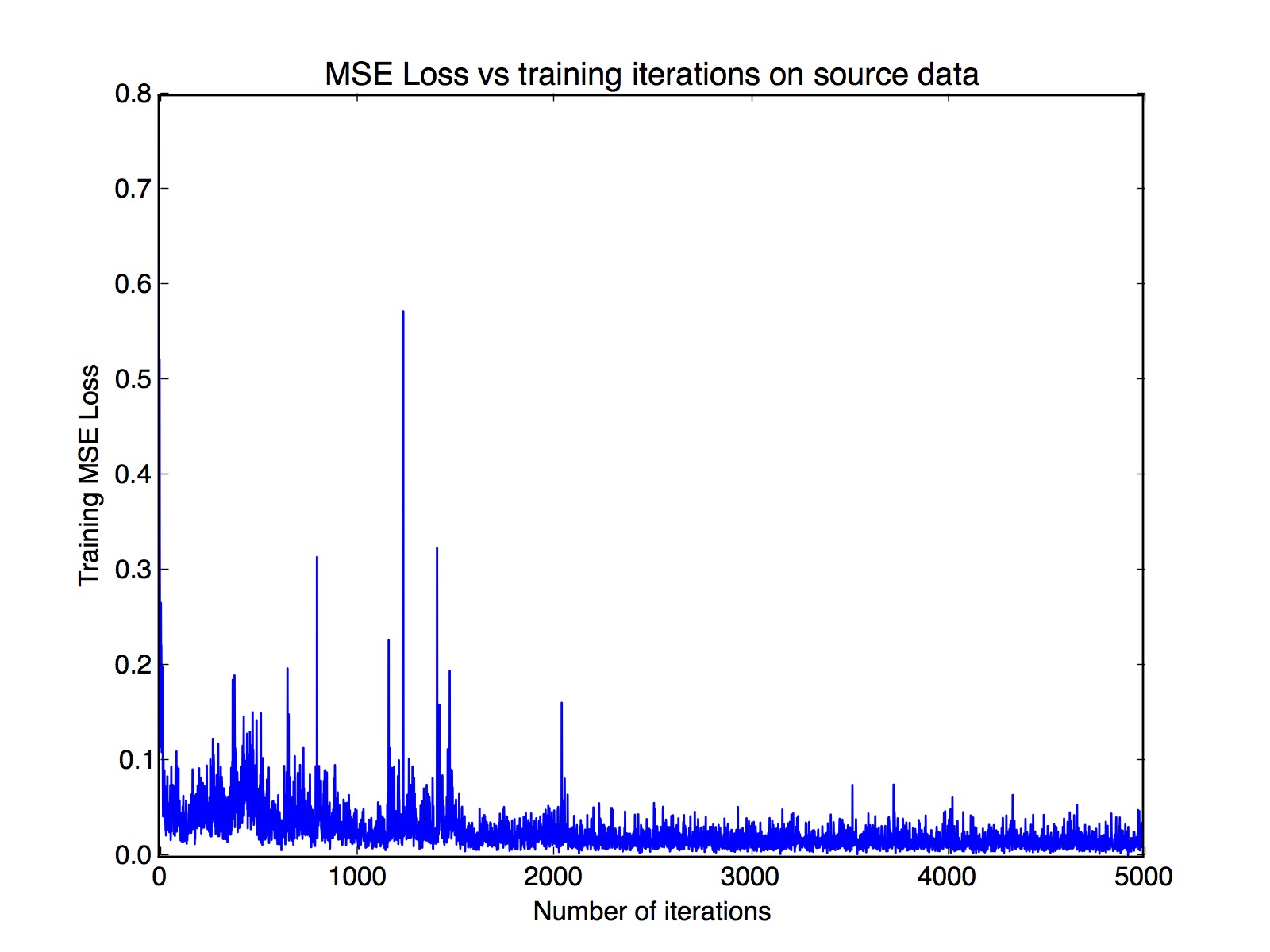}
\includegraphics[width=0.4\textwidth]{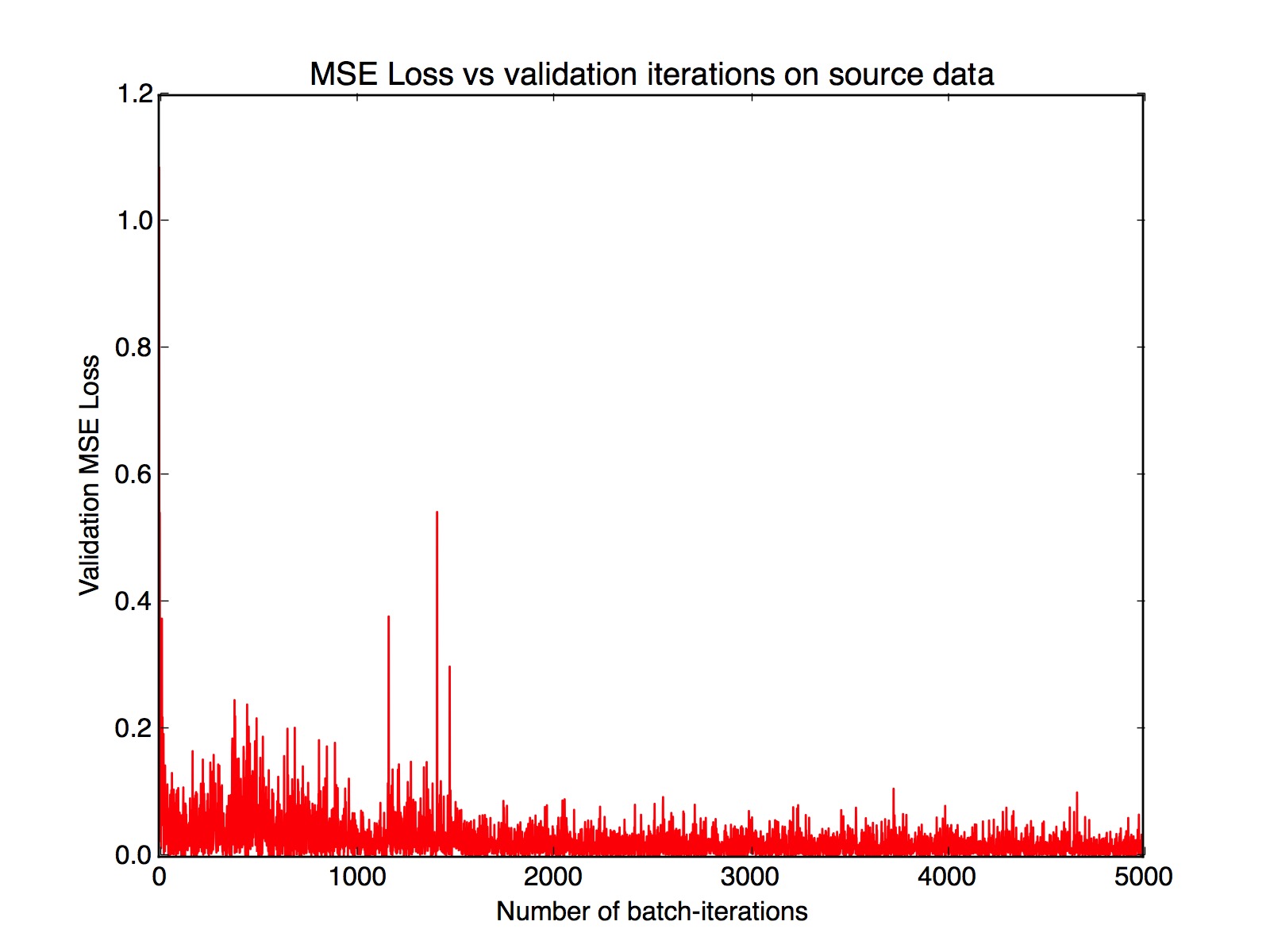}
\caption{The plot that shows the variation of training MSE loss(left) and validation MSE loss(right) with respect to number of iterations}
\end{figure}
The loss values on the training data started from 0.75 and after 5000 iterations with a batch-size of 32 ended up being 0.02. Also on the validation set a similar trend was observed where the validation loss went from 1.2 radians average to 0.04.

\subsection{Results for phase 2}
{\bf Results for Hypothesis 1: } \\
We also found that leaky ReLu performs better than ReLu as ReLu renders many gradients to 0 which makes it harder to train the generator network. 

\begin{figure}
\centering
\begin{minipage}{0.5\textwidth}
  \centering
  \includegraphics[width=0.75\linewidth]{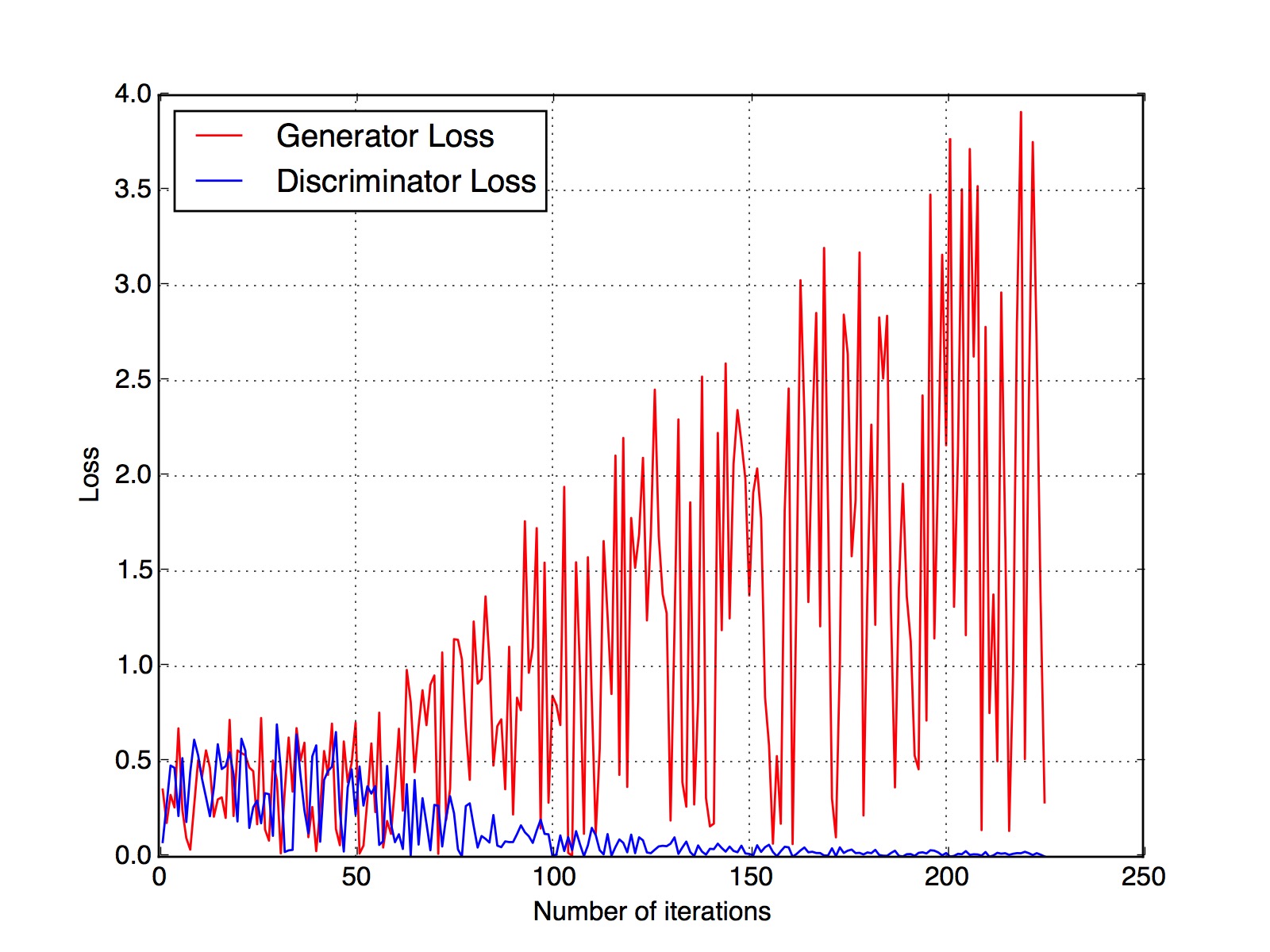}
\end{minipage}%
\begin{minipage}{.5\textwidth}
  \centering
  \includegraphics[width=0.75\linewidth]{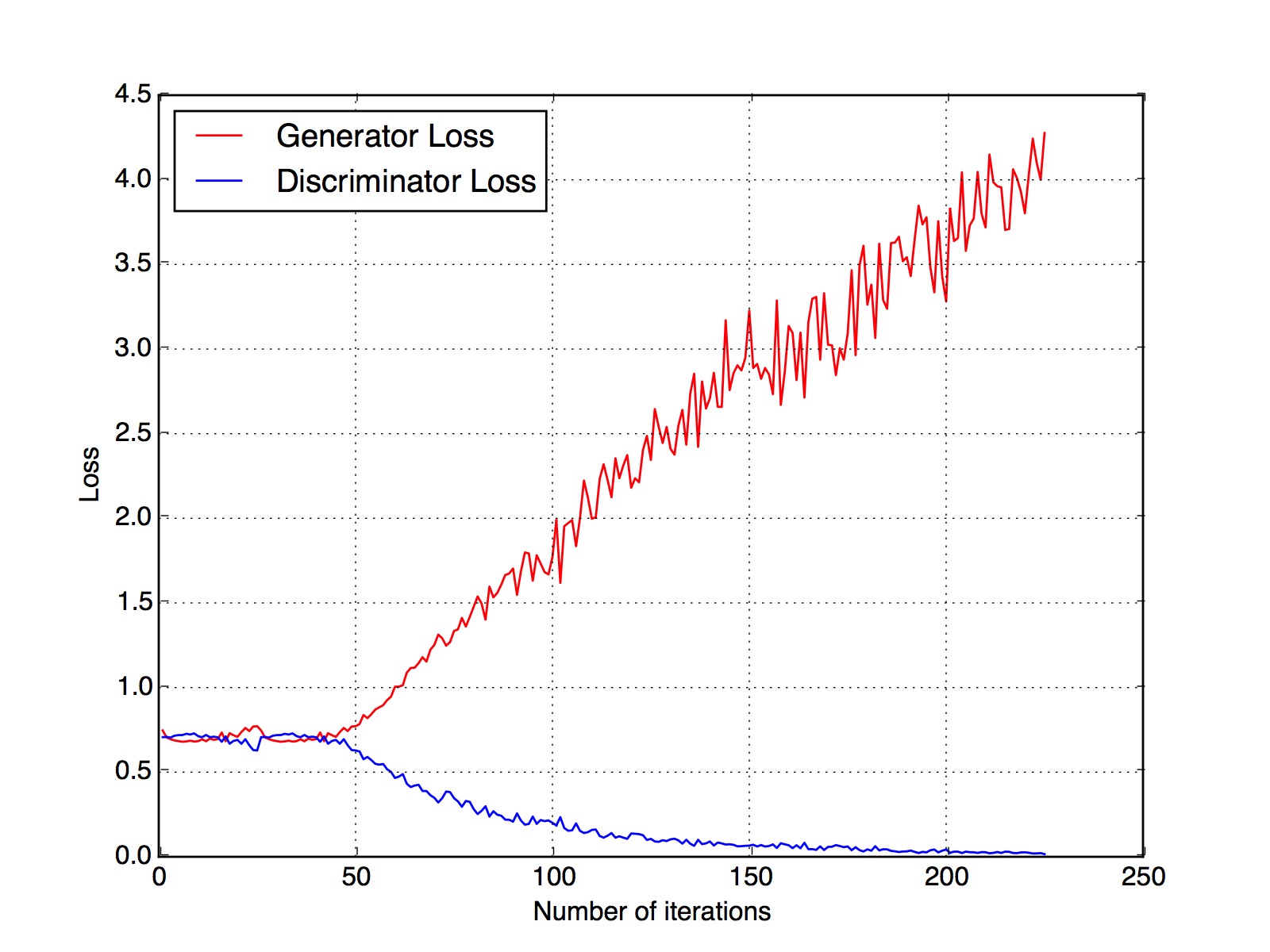}
\end{minipage}
\caption{Plots testing Hypothesis 1. a. Plots of $L_{S\Leftarrow T}^{G}$ and $L_{S\Leftarrow T}^{D}$ using Relu activations, b. Plots of $L_{S\Leftarrow T}^{G}$ and $L_{S\Leftarrow T}^{D}$ using leaky Relu activations}
\label{fig:hyp1}
\end{figure}

The results are apparent in Figure \ref{fig:hyp1} in which we find that ReLu training is very unstable while leaky ReLu training achieves more stability.

\begin{figure}
\centering
\begin{minipage}{0.5\textwidth}
  \centering
  \includegraphics[width=0.75\linewidth]{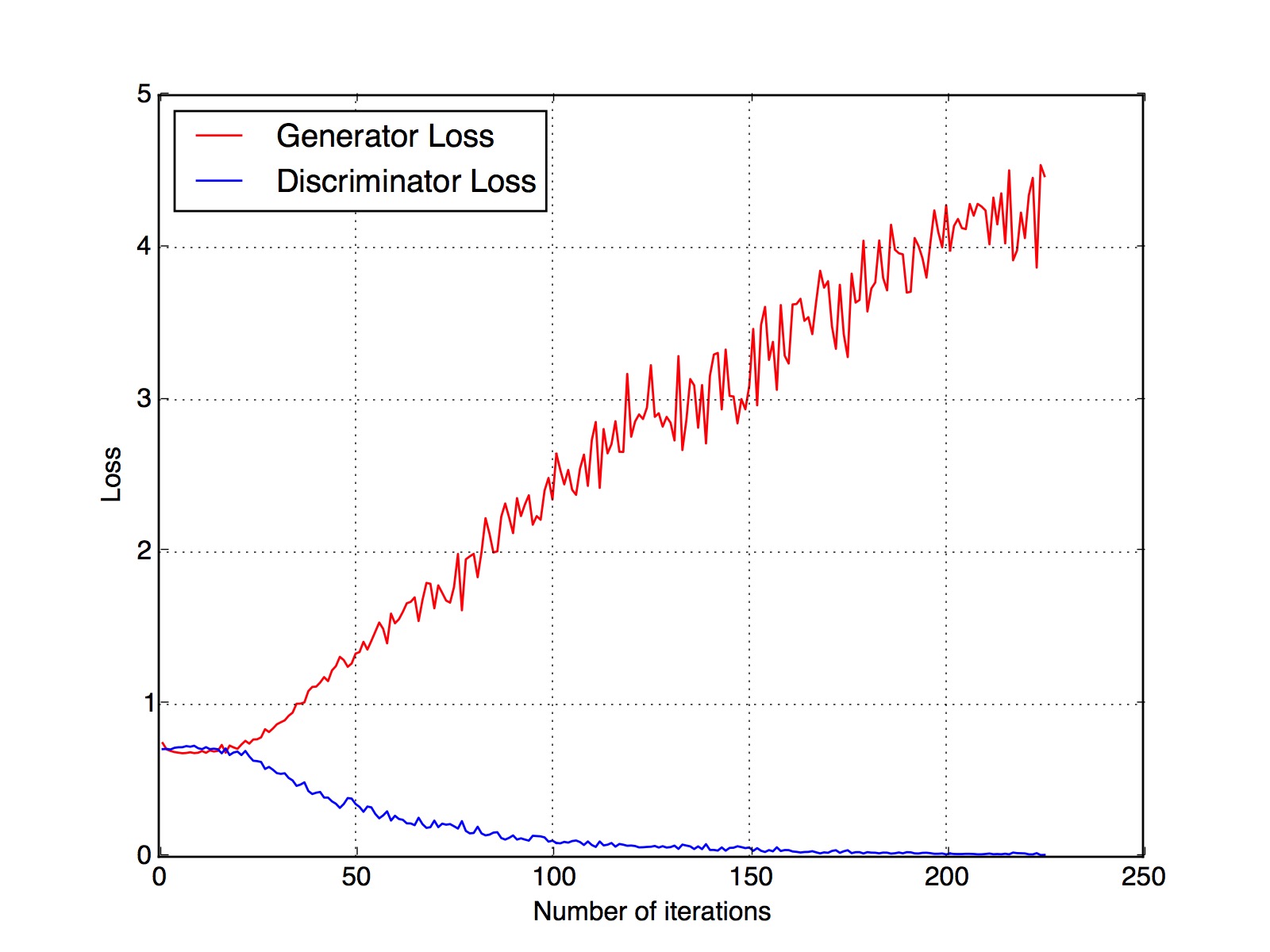}
\end{minipage}%
\begin{minipage}{.5\textwidth}
  \centering
  \includegraphics[width=0.75\linewidth]{minimize.jpg}
\end{minipage}
\caption{Plots for testing Hypothesis 2. a. Plots of $L_{S\rightarrow T}^{G}$ and $L_{S\rightarrow T}^{D}$  for  generator objective which maximizes discriminator loss. 
 b. Plot of $L_{S\rightarrow T}^{G}$ and $L_{S\rightarrow T}^{D}$ for generator objective which minimizes flipped discriminator logits
}
\label{fig:hyp2}
\end{figure}

{\bf Results for Hypothesis 2: } \\
We find experimentally that minimizing flipped logits of discriminator leads to better convergence than maximizing discriminator loss. We can see the plots in Figure \ref{fig:hyp2} in which the first plot starts to diverge early while the second plot diverges after a new iterations.

\begin{figure}
\centering
\begin{minipage}{0.5\textwidth}
  \centering
  \includegraphics[width=0.75\linewidth]{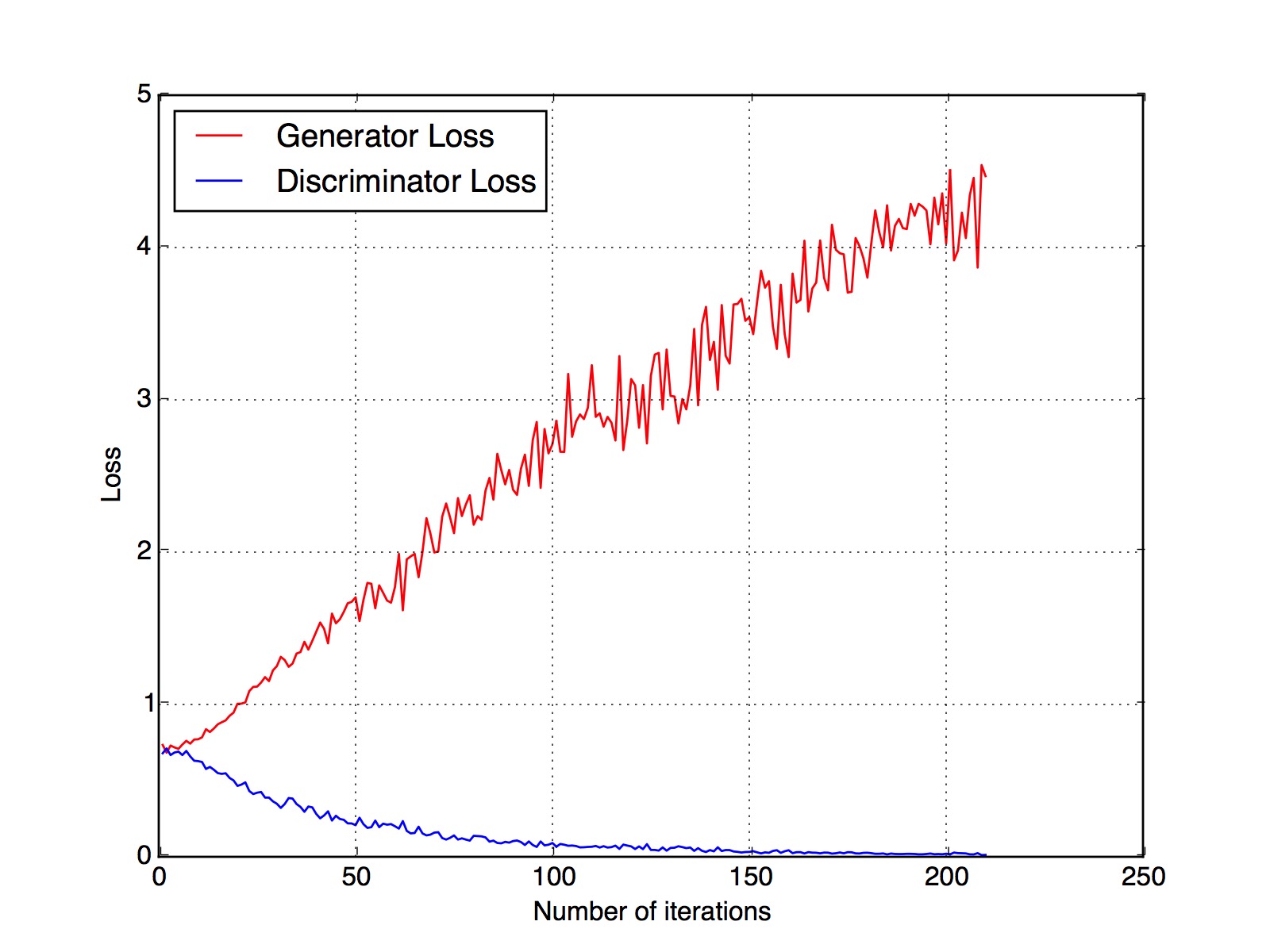}
\end{minipage}%
\begin{minipage}{.5\textwidth}
  \centering
  \includegraphics[width=0.75\linewidth]{minimize.jpg}
\end{minipage}
\caption{Plots for testing Hypothesis 3. a. $L_{S\rightarrow T}^{G}$ and $L_{S\rightarrow T}^{D}$  for shuffled source and target data in batches during training.
 b. Plot of $L_{S\rightarrow T}^{G}$ and $L_{S\rightarrow T}^{D}$ for separate mini-batches for source and target data during training.
}
\label{fig:hyp3}
\end{figure}

{\bf Results for Hypothesis 3: } \\
From the plots in Figure \ref{fig:hyp3}, it is apparent that we achieve better results with separate mini-batches. The generator loss $L_{S\Leftarrow T}^{G}$ starts to diverge at approximately 50 iterations in the separate mini-batch training method. The loss function in shuffled mini-batches diverges very early on.

\begin{figure}
\centering
\begin{minipage}{0.5\textwidth}
  \centering
  \includegraphics[width=0.75\linewidth]{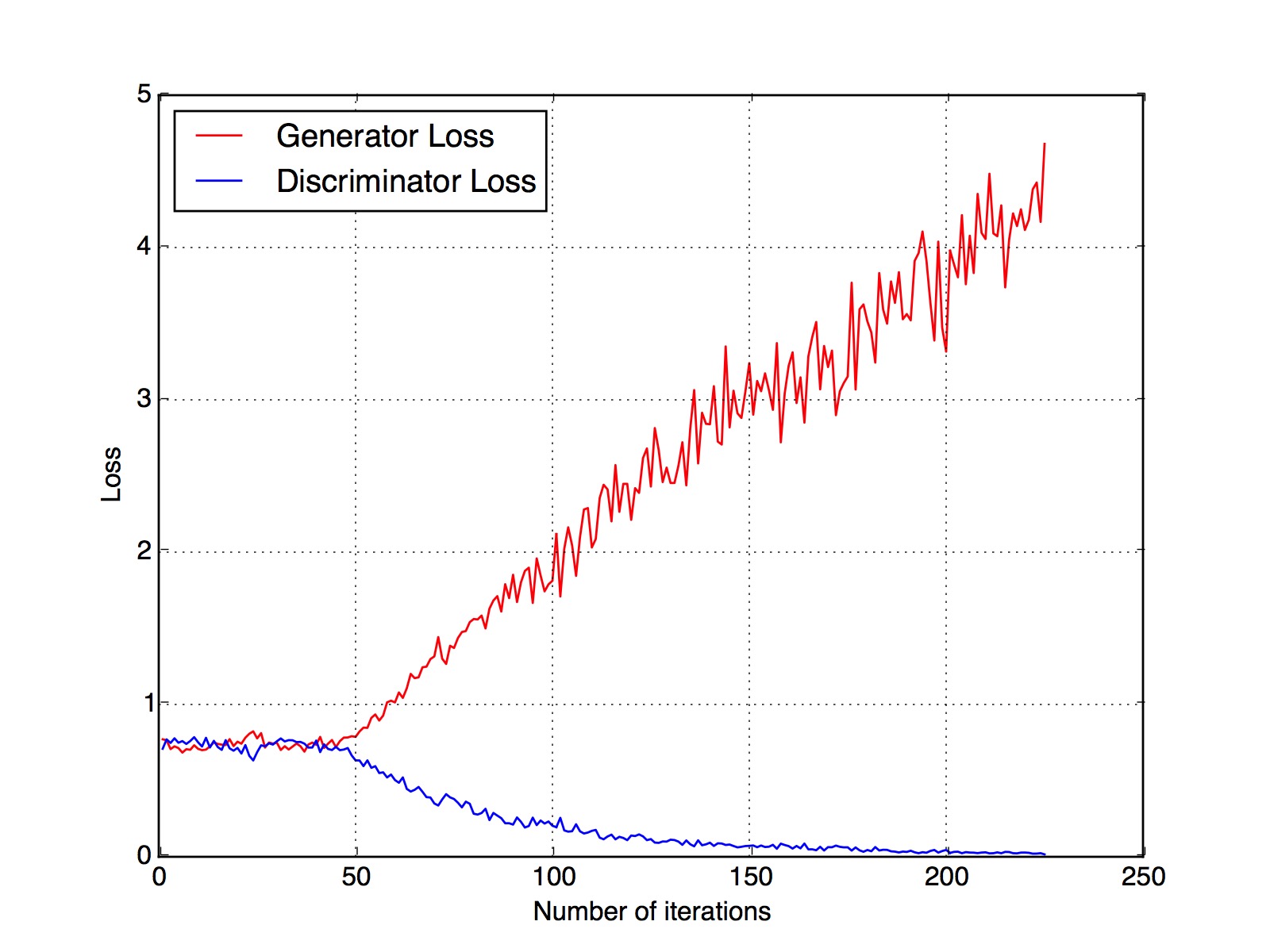}
\end{minipage}%
\begin{minipage}{.5\textwidth}
  \centering
  \includegraphics[width=0.75\linewidth]{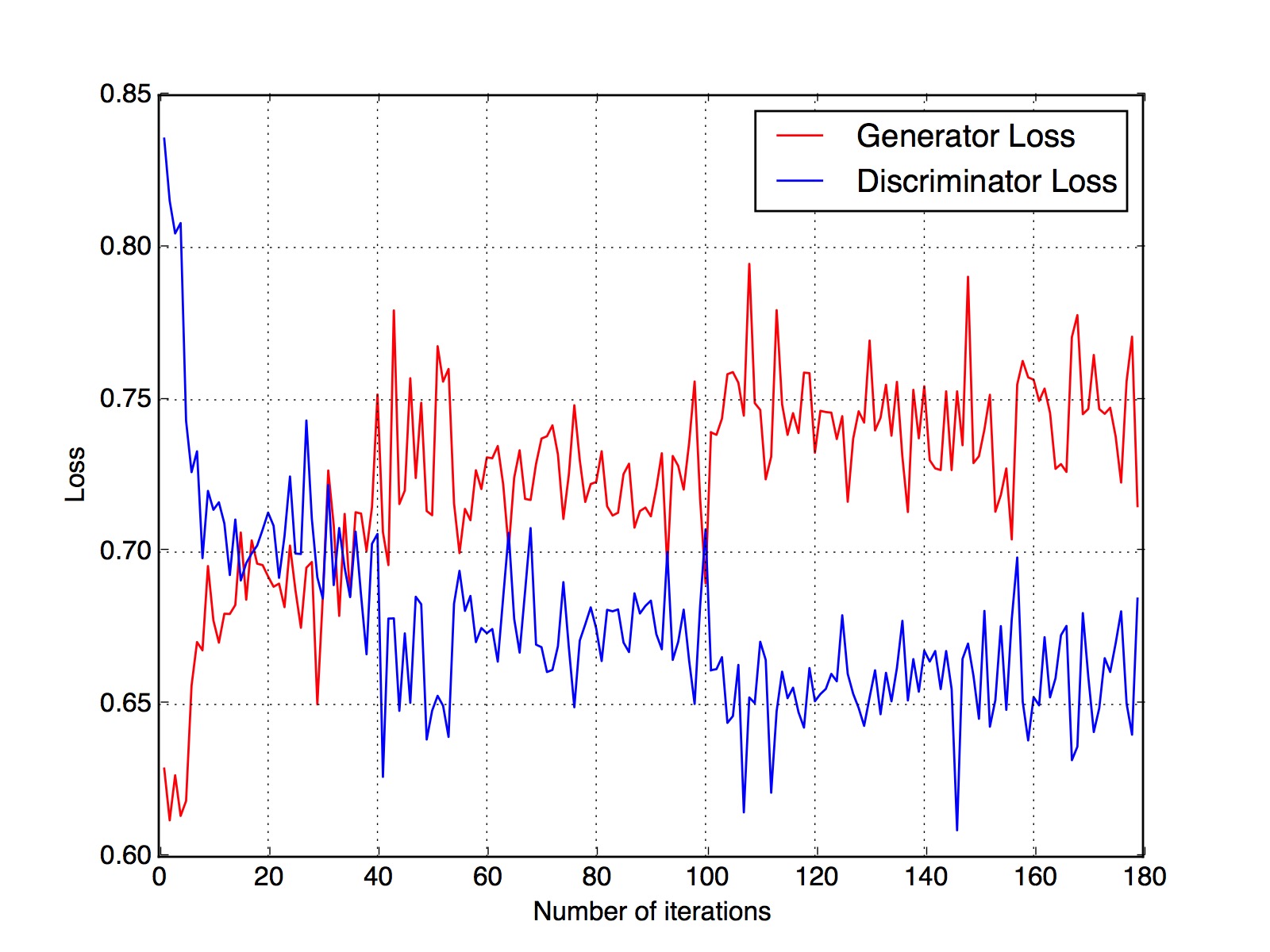}
\end{minipage}
\caption{Plots for testing Hypothesis 4. a. Plots of $L_{S\rightarrow T}^{G}$ and $L_{S\rightarrow T}^{D}$  for balanced training.
 b. Plot of $L_{S\rightarrow T}^{G}$ and $L_{S\rightarrow T}^{D}$ for sparse training.
}
\label{fig:hyp4}
\end{figure}

{\bf Results for Hypothesis 4: } \\
Experimentally, we obtained better results for sparse training than balanced training.
We tried 2 threshold values: 0.7 and 0.8. We found that 0.8 is better for training than 0.7. Discriminator loss of 0.7 corresponds to discriminator logits of 0.5 ($log(0.5)=-0.69$). This means that the discriminator predicts that the data belongs to target or source domain with 0.5 probability. This is also the equilibrium value for a successfully trained GAN network \citep{goodfellow2014generative}. Hence, its a very strict threshold to keep as the training gets stuck in the while loop of generator. The results for 0.8 threshold value as can be seen in the plot of Figure \ref{fig:hyp4} b.

{\bf Results for Hypothesis 5: } \\ 
In order to learn a generic regressor, it is essential to produce images that appear to belong to source or target distribution as closely as possible. If we are able to achieve images of better quality, we can simply plug the source regressor on target images by transforming target to source image and then using trained source regressor. While we believe that with more computational resources and time, we can generate good images using GANs, we also wanted to introduce a new loss function which directly ensures valid transformation from source to target and target to source. This loss is the reconstruction loss as described in Section 4. Using this loss, we trained our network for 200 iterations and we show how this loss augments GAN training in the results section.

The GAN network produces images which retain semantic information from the input image like the roads, trees, etc. The roads also retain the angle of the input image. Following images are produced by adversarial learning of domain transfer networks. \\

\begin{figure}
\centering
\begin{minipage}{0.5\textwidth}
  \centering
  \includegraphics[width=0.75\linewidth]{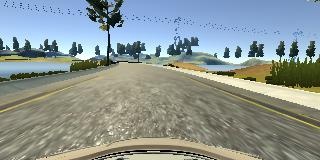}
\end{minipage}%
\begin{minipage}{.5\textwidth}
  \centering
  \includegraphics[width=0.75\linewidth]{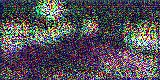}
\end{minipage}

\centering
\begin{minipage}{0.5\textwidth}
  \centering
  \includegraphics[width=0.75\linewidth]{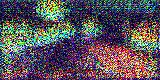}
\end{minipage}%
\begin{minipage}{.5\textwidth}
  \centering
  \includegraphics[width=0.75\linewidth]{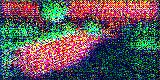}
\end{minipage}

\caption{Plots testing Hypothesis 5. Networks are trained using only adversarial loss, (images labeled clockwise from top-left corner) a. Source image, b. Image generated by source to target generator network trained for 100 iterations, c. Image generated by source to target generator network trained for 150 iterations, d. Image generated by source to target generator network trained for 200 iterations}

\label{fig:s2t}
\end{figure}

\begin{figure}
\centering
\begin{minipage}{0.5\textwidth}
  \centering
  \includegraphics[width=0.75\linewidth]{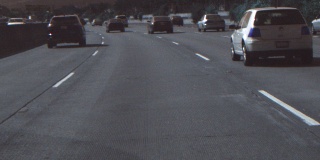}
\end{minipage}%
\begin{minipage}{.5\textwidth}
  \centering
  \includegraphics[width=0.75\linewidth]{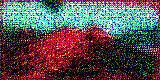}
\end{minipage}

\centering
\begin{minipage}{0.5\textwidth}
  \centering
  \includegraphics[width=0.75\linewidth]{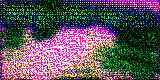}
\end{minipage}%
\begin{minipage}{.5\textwidth}
  \centering
  \includegraphics[width=0.75\linewidth]{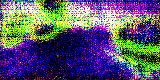}

\label{fig:t2s}
\end{minipage}

\caption{Plots testing Hypothesis 5. Networks are trained using only adversarial loss, (images labeled clockwise from top-left corner) a. Target image, b. Image generated by target to source generator network trained for 100 iterations, c. Image generated by target to source generator network trained for 150 iterations, d. Image generated by target to source generator network trained for 200 iterations}
\end{figure}

\begin{figure}
\centering
\begin{minipage}{0.5\textwidth}
  \centering
  \includegraphics[width=0.75\linewidth]{source.jpg}
\end{minipage}%
\begin{minipage}{.5\textwidth}
  \centering
  \includegraphics[width=0.75\linewidth]{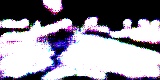}
\end{minipage}

\centering
\begin{minipage}{0.5\textwidth}
  \centering
  \includegraphics[width=0.75\linewidth]{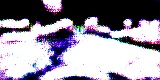}
\end{minipage}%
\begin{minipage}{.5\textwidth}
  \centering
  \includegraphics[width=0.75\linewidth]{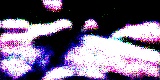}
\end{minipage}

\caption{Plots for source to target generator learnt using adversarial and reconstruction loss, (images labeled clockwise from top-left corner) a. Source image, b. Source image converted to a target image using a source to target generator network trained for 100 epochs, c. Source image converted to a target image using a source to target generator network trained for 150 epochs, d. Source image converted to a target image using a source to target generator network trained for 200 epochs}
\label{fig:recon}
\end{figure}

As we can see from generated target images (Figure \ref{fig:s2t}), the scene doesn't appear to be real world as we had hoped. Same can be said about generated source images from target image in (Figure \ref{fig:t2s}). In Figure \ref{fig:recon}, we were able to produce images of closer correlation with the image than before.

Note that the generated images are still not the same as the actual images from source or target dataset. We attribute the reason to this as the lack of resources to train a big network. With more time and compute power, we think we will be able to achieve better results for generation.

Hence, we use composite loss of GAN loss and reconstruction loss in Phase 2. The weights learnt in this Phase are used to initialize the training network of Phase 3. Note that these weights are continued to be learnt and updated in Phase 3.

\subsection{Phase 3}
\begin{figure}[htbp]
\centering
\includegraphics[width=0.7\textwidth]{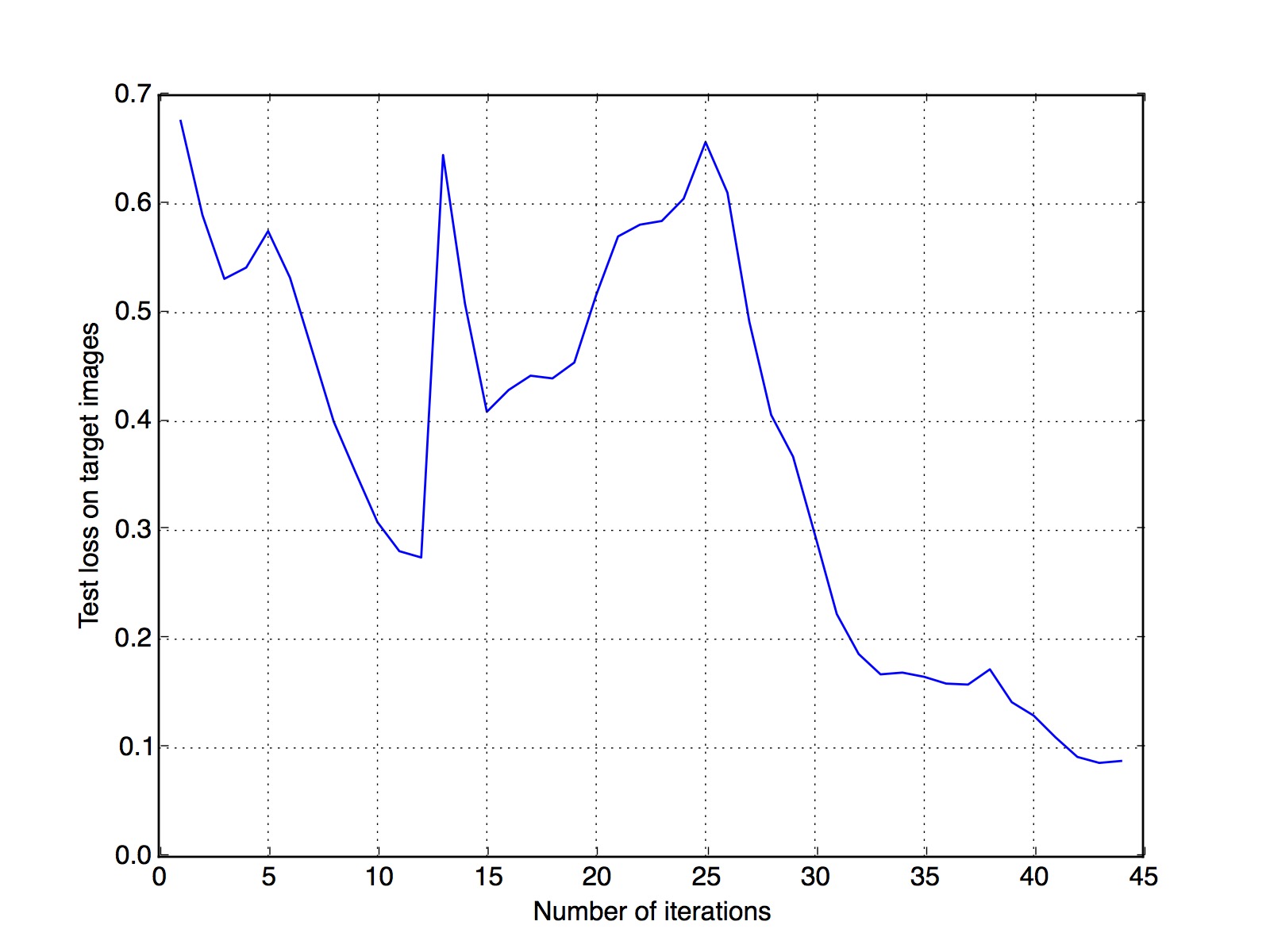}
\caption{Validation loss during training for 430 iterations}
\label{fig:test}
\end{figure}

In Figure \ref{fig:test} is the plot of validation loss of a target video of 1000 frames. The video sequence was passed through the regressor at every tenth iteration of training. We plotted the loss of regressor on validation set as the regressor progressed through Phase 3 training. The loss decreases almost consistently from 0.687 to 0.082. While this loss is not as low as the training loss on source images which was 0.02, we certainly feel that this method has potential to perform better given more time and compute resources. We also report final test MSE loss for a held-out video of 1000 frames. This loss was value was 0.091. We compare the performance of our training method with directly using source regressor trained in the first step. Hence, the performance of source regressor from the first step on validation and test sets is our baseline. We report an improvement of 12.09\% of Average Absolute Relative Error (AARE) on test data with our proposed training method. The average MSE value decreases by 0.139. 

\begin{table}[t]
  \centering
  \begin{tabular}{|l|l|l|l|l|}
    \toprule

 & Val MSELoss  & Val AARE & Test MSELoss & Test AARE \\
 \hline
Adve trained regressor & 0.082 & 27.33\% & 0.091 &  31.43\% \\
Source regressor & 0.15 &  48.66\% & 0.23 & 43.52\% \\
    \bottomrule
  \end{tabular}
  \vspace{1mm}
  \caption{Final Validation and Test loss on held out datasets}
  \label{test}
\end{table}

\section{Discussion and Conclusions}
We refer to the final graph of Experiments section. Note that a loss 0.09 means that the steering angle could be as wrong as 0.09 radians on an average. To put this to perspective, 0.09 radians is approximately 4.5 degrees which could prove fatal if errors compound. Hence, our system cannot be deployed in any way but our preliminary results show that this method of generalization and domain adaptation is promising. \\

In this work, we came up with several novel ideas to constrain the GAN training problem in a way which makes its learning more stable and have tighter semantic regeneration goals. While rudimentary, our experiments do point towards a possibly working direction and certain new ideas which can be tested based on what our results showed. Our generator networks, despite being shallow by any real world standard, were successfully able to capture semantic information from the target domain images. Even though the synthesized images were not photo-realistic, as was expected by the Loss function that we tried to constrain all the networks with, the semantic information retention is a positive outcome from our experiments. \\

We have some untested hypotheses and ideas for improving the quality of synthetic images (which would help us solve the auxiliary goal of generating labeled data without supervision). Firstly, we can have deeper convolutional networks, which would allow the networks to learn more semantic features. Secondly, we can add skip connections in intermediate high level feature layers, preferably feeding the output of a convolution unit in the first half to the input of a deconvolution unit in the second half. Thirdly, we could, in principle, use pre-trained networks which are more adept at image regeneration by training an auto-encoder on both target and source domain, and then use them to create the initial states for domain translation networks. Such a scheme could be realized by plugging in some slim layers between, say, the encoder half of source domain auto-encoder, and the second half of say the target domain auto-encoder. After training the slim intermediate layers for some iterations by keeping the other parts frozen, we could train the full network, which, we expect would have already learnt image regeneration to a higher extent than what we start with in our first phase of training.\\

The idea of using helper networks to augment GAN training has several exciting extensions in terms of making GAN learning robust in general. With semantic reconstruction loss, we were better able to constrain the Generator Discriminator networks to learn in a more stable manner. Similarly, using tweaks based on sparse learning for Discriminator also increased the robustness of learning a lot. A very direct future application is to use a unified Discriminator for both the cross domain translator networks. Also, we could experiment with a steering angle regression network which solely works on the target data, to bring about a total symmetry in our phase 3 training.\\

\bibliographystyle{unsrtnat}
\bibliographystyle{spbasic}      
\bibliography{bibliography}

\begin{thebibliography}{11}
\providecommand{\natexlab}[1]{#1}
\providecommand{\url}[1]{\texttt{#1}}
\expandafter\ifx\csname urlstyle\endcsname\relax
  \providecommand{\doi}[1]{doi: #1}\else
  \providecommand{\doi}{doi: \begingroup \urlstyle{rm}\Url}\fi

\bibitem[Prettenhofer and Stein(2010)]{prettenhofer2010cross}
Peter Prettenhofer and Benno Stein.
\newblock Cross-language text classification using structural correspondence
  learning.
\newblock In \emph{Proceedings of the 48th annual meeting of the association
  for computational linguistics}, pages 1118--1127. Association for
  Computational Linguistics, 2010.

\bibitem[Gretton et~al.(2009)Gretton, Smola, Huang, Schmittfull, Borgwardt, and
  Sch{\"o}lkopf]{gretton2009covariate}
Arthur Gretton, Alexander~J Smola, Jiayuan Huang, Marcel Schmittfull, Karsten~M
  Borgwardt, and Bernhard Sch{\"o}lkopf.
\newblock Covariate shift by kernel mean matching.
\newblock 2009.

\bibitem[Tzeng et~al.(2014)Tzeng, Hoffman, Zhang, Saenko, and
  Darrell]{tzeng2014deep}
Eric Tzeng, Judy Hoffman, Ning Zhang, Kate Saenko, and Trevor Darrell.
\newblock Deep domain confusion: Maximizing for domain invariance.
\newblock \emph{arXiv preprint arXiv:1412.3474}, 2014.

\bibitem[Long et~al.(2015)Long, Cao, Wang, and Jordan]{long2015learning}
Mingsheng Long, Yue Cao, Jianmin Wang, and Michael Jordan.
\newblock Learning transferable features with deep adaptation networks.
\newblock In \emph{International Conference on Machine Learning}, pages
  97--105, 2015.

\bibitem[Tzeng et~al.(2015)Tzeng, Hoffman, Darrell, and
  Saenko]{tzeng2015simultaneous}
Eric Tzeng, Judy Hoffman, Trevor Darrell, and Kate Saenko.
\newblock Simultaneous deep transfer across domains and tasks.
\newblock In \emph{Proceedings of the IEEE International Conference on Computer
  Vision}, pages 4068--4076, 2015.

\bibitem[Ganin and Lempitsky(2015)]{ganin2015unsupervised}
Yaroslav Ganin and Victor Lempitsky.
\newblock Unsupervised domain adaptation by backpropagation.
\newblock In \emph{International Conference on Machine Learning}, pages
  1180--1189, 2015.

\bibitem[Goodfellow et~al.(2014)Goodfellow, Pouget-Abadie, Mirza, Xu,
  Warde-Farley, Ozair, Courville, and Bengio]{goodfellow2014generative}
Ian Goodfellow, Jean Pouget-Abadie, Mehdi Mirza, Bing Xu, David Warde-Farley,
  Sherjil Ozair, Aaron Courville, and Yoshua Bengio.
\newblock Generative adversarial nets.
\newblock In \emph{Advances in neural information processing systems}, pages
  2672--2680, 2014.

\bibitem[Tzeng et~al.(2017)Tzeng, Hoffman, Saenko, and
  Darrell]{tzeng2017adversarial}
Eric Tzeng, Judy Hoffman, Kate Saenko, and Trevor Darrell.
\newblock Adversarial discriminative domain adaptation.
\newblock \emph{arXiv preprint arXiv:1702.05464}, 2017.

\bibitem[Hoffman et~al.(2018)Hoffman, Tzeng, Park, Zhu, Isola, Saenko, Efros,
  and Darrell]{hoffman2018cycada}
Judy Hoffman, Eric Tzeng, Taesung Park, Jun-Yan Zhu, Phillip Isola, Kate
  Saenko, Alexei Efros, and Trevor Darrell.
\newblock Cycada: Cycle-consistent adversarial domain adaptation.
\newblock In \emph{International conference on machine learning}, pages
  1989--1998. Pmlr, 2018.

\bibitem[Xiang and Li(2017)]{norm}
Sitao Xiang and Hao Li.
\newblock On the effects of batch and weight normalization in generative
  adversarial networks.
\newblock \emph{stat}, 1050:\penalty0 22, 2017.

\bibitem[Radford et~al.(2015)Radford, Metz, and Chintala]{DCGAN}
Alec Radford, Luke Metz, and Soumith Chintala.
\newblock Unsupervised representation learning with deep convolutional
  generative adversarial networks.
\newblock \emph{arXiv preprint arXiv:1511.06434}, 2015.

\end{thebibliography}

\end{document}